\documentclass{article}

\usepackage{arxiv}

\usepackage[utf8]{inputenc} 
\usepackage[T1]{fontenc}    
\usepackage{hyperref}       
\usepackage{url}            
\usepackage{booktabs}       
\usepackage{amsfonts}       
\usepackage{amsmath}
\usepackage{listings}
\usepackage{fancyvrb}
\usepackage{tabularx}
\usepackage{nicefrac}       
\usepackage{microtype}      
\usepackage{lipsum}		
\usepackage{graphicx}
\usepackage{natbib}
\usepackage{doi}

\title{SNeL: A Structured Neuro-Symbolic Language for Entity-Based Multimodal Scene Understanding}

\author{
	\href{https://orcid.org/0000-0001-6269-494X}{\includegraphics[scale=0.06]{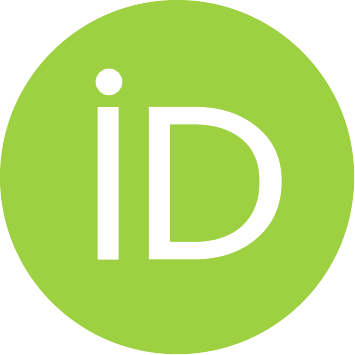}\hspace{1mm}Silvan Ferreira} \\
	Federal University of Rio Grande do Norte  \\
	Natal, Brazil \\
	\texttt{silvan.junior.051@ufrn.edu.br} \\
	\And
	\href{https://orcid.org/0000-0002-9486-4509}{\includegraphics[scale=0.06]{orcid.pdf}\hspace{1mm}Allan Martins} \\
	Federal University of Rio Grande do Norte \\
	Natal, Brazil \\
	\texttt{allan@dee.ufrn.br} \\
	\And
	\href{https://orcid.org/0000-0002-0116-6489}{\includegraphics[scale=0.06]{orcid.pdf}\hspace{1mm}Ivanovitch Silva} \\
	Federal University of Rio Grande do Norte \\
	Natal, Brazil \\
	\texttt{ivanovitch.silva@ufrn.br} \\
}

\date{}

\renewcommand{\shorttitle}{SNeL: Structured Neuro-symbolic Language}

\hypersetup{
pdftitle={SNeL: A Structured Neuro-Symbolic Language for Entity-Based Multimodal Scene Understanding},
pdfsubject={cs.AI},
pdfauthor={Silvan Ferreira, Allan Martins, Ivanovitch Silva},
pdfkeywords={Neural Networks, Query Language, Scene Understanding, Entity Detection, Multimodal Deep Learning, Neuro-Symbolic AI},
}

\begin{document}
\maketitle

\begin{abstract}
In the evolving landscape of artificial intelligence, multimodal and Neuro-Symbolic paradigms stand at the forefront, with a particular emphasis on the identification and interaction with entities and their relations across diverse modalities. Addressing the need for complex querying and interaction in this context, we introduce SNeL (Structured Neuro-symbolic Language), a versatile query language designed to facilitate nuanced interactions with neural networks processing multimodal data. SNeL's expressive interface enables the construction of intricate queries, supporting logical and arithmetic operators, comparators, nesting, and more. This allows users to target specific entities, specify their properties, and limit results, thereby efficiently extracting information from a scene. By aligning high-level symbolic reasoning with low-level neural processing, SNeL effectively bridges the Neuro-Symbolic divide. The language's versatility extends to a variety of data types, including images, audio, and text, making it a powerful tool for multimodal scene understanding. Our evaluations demonstrate SNeL's potential to reshape the way we interact with complex neural networks, underscoring its efficacy in driving targeted information extraction and facilitating a deeper understanding of the rich semantics encapsulated in multimodal AI models.
\end{abstract}


\section{Introduction}
The rapidly evolving field of artificial intelligence (AI) has seen a significant shift toward more complex, holistic, and integrated forms of understanding. The rise of multimodal AI, capable of processing and integrating information across various modalities such as images, audio, and text, has unlocked new possibilities for more sophisticated and nuanced system interactions. Concurrently, the Neuro-Symbolic AI paradigm has emerged, combining the power of neural networks with the interpretability of symbolic reasoning, thereby bridging the gap between high-level reasoning and low-level data processing.

\begin{figure}[t!]
  \centering
  \includegraphics[width=1.0\linewidth]{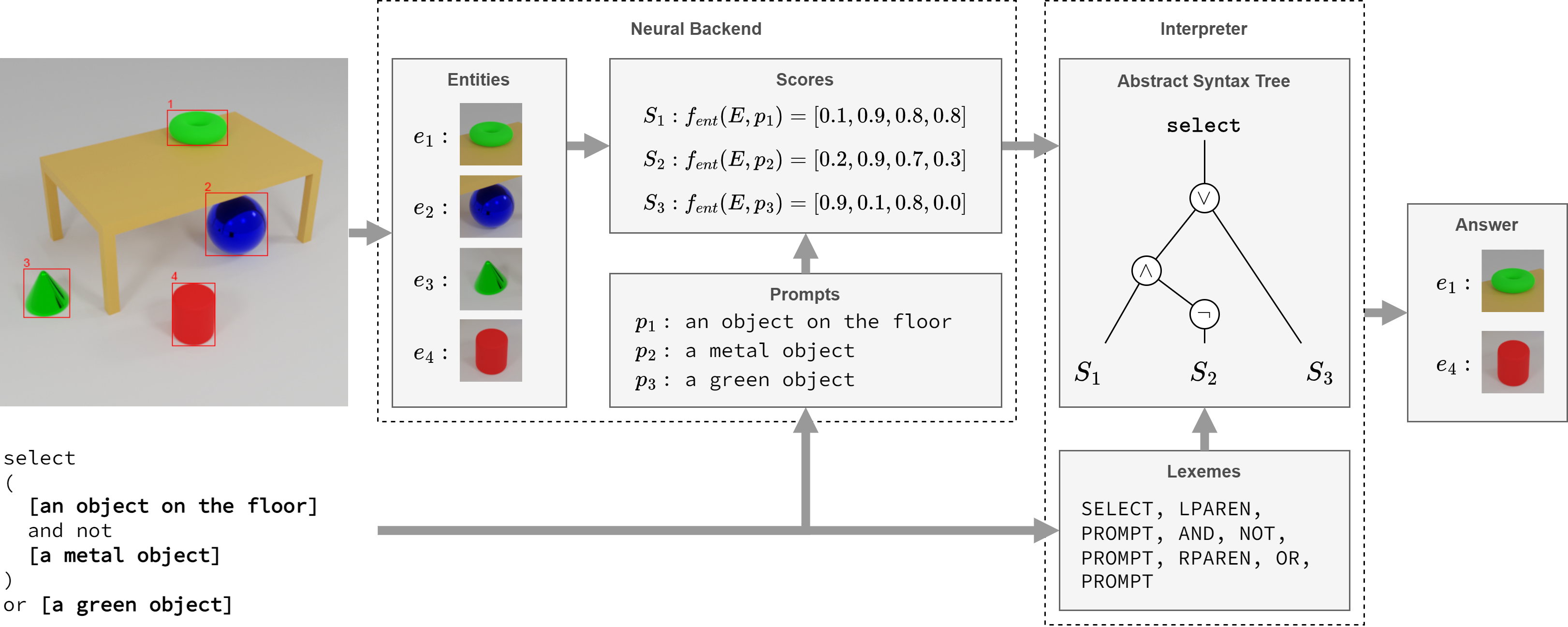}
  \caption{General architecture of SNeL.}
  \label{fig:snel}
\end{figure}

In the complex and highly semantic field of AI, the ability to recognize and engage with elements and their connections within a context is crucial. Entities, as the basic units of perception and understanding, form the cornerstone of our ability to make sense of complex environments. However, efficiently accessing, manipulating, and querying these entities within neural network models poses a significant challenge. The conventional methods of interaction with these models often lack the necessary granularity and flexibility, failing to fully exploit the rich semantics encapsulated within them. Therefore, there is a compelling need for a refined, more expressive language that can facilitate nuanced interactions with neural networks, particularly those handling multimodal data. Such a language should enable users to construct complex queries, targeting specific entities, specifying their properties, and limiting results, thereby allowing for targeted information extraction and in-depth scene understanding. This forms the primary motivation for our work, paving the way for the development of SNeL - the Structured Neuro-symbolic Language, designed to bring together the strengths of Neuro-Symbolic AI and multimodal deep learning. The general architecture of the proposed language can be seen in Figure \ref{fig:snel}.

The proposed system is fundamentally grounded on the ontological concept of entities, which serve as the fundamental units of perception and comprehension across different modalities. This allows for the semantic range to encompass the use of complex queries, with the use of logical and arithmetic operators, comparators, and the capacity for query grouping. Additionally, the entity-centered approach extends to many modalities, including image, video, audio and text, thereby establishing it as a truly multimodal language. This versatility is facilitated by the integration of deep learning models that extract and score entities based on input text queries. These models process the data, identify potential entities, and evaluate their relevance based on the alignment between the query and the identified entities.


Our contributions in this paper are anchored in the design and implementation of SNeL, a novel language purpose-built to facilitate nuanced and sophisticated interactions with multimodal neural networks in the context of scene understanding and entity detection. The proposed query language introduces an expressive interface that enables users to employ text prompts to construct complex queries targeting specific entities within a scene across different modalities. By aligning high-level symbolic reasoning with low-level neural processing, SNeL effectively bridges the Neuro-Symbolic divide and presents a novel tool for advanced AI scene understanding.

The structure of this paper is organized as follows: In Related Work, we explore existing literature on multimodal and Neuro-Symbolic AI, entity detection, and query languages. This is followed by the Language Design Section, where we detail SNeL's syntax, semantics, support for complex queries and the clauses used to assembly queries. Next, in Language Components, we describe the integration of the proposed language with underlying neural networks. Then, in Use Cases and Examples Section, we provide examples of applications across multiple domains. Finally, the paper concludes with a summary of limitations, potential enhancements, and future research directions.

\section{Related Work}
The development of the SNeL system can be best understood in the context of substantial prior work across several interrelated fields. This encompasses methodologies for entity detection across multiple domains such as Computer Vision, Natural Language Processing, and Audio Processing, advances in cross-modal alignment and information retrieval and the application of context-free languages (CFL) in Neuro-Symbolic AI. Each of these components has contributed to the formation of SNeL's architecture, which effectively integrates these varying facets. In this Section, we provide an overview of the key contributions and methodologies in these areas that have influenced the design and capabilities of the proposed system.

\subsection{Entity Detection}

In the field of Computer Vision, entities can be distinct objects within an image or a video scene. Here, Entity Recognition can be associated with tasks like object detection, where the aim is to identify specific objects within a given image and determine their boundaries \citep{zou2023object}. Notable work in this domain includes the R-CNN family of models, including the original R-CNN \citep{girshick2014rich}, Fast R-CNN \citep{girshick2015fast}, and Faster R-CNN \citep{ren2015faster}. These models employ an approach known as Region Proposal Networks (RPNs) to suggest potential object locations, before classifying and refining these locations with a separate network. Another significant advancement in the field is the You Only Look Once (YOLO) algorithm \citep{redmon2016you}. In stark contrast to the region proposal methodology, YOLO uses a single Convolutional Neural Network (CNN) that simultaneously predicts multiple bounding boxes and class probabilities for those boxes.

In Natural Language Processing, Text segmentation is a technique that partitions a document into smaller, meaningful units, often termed as "segments". Segments could be delineated based on a variety of factors such as words, sentences, topics, phrases, or any relevant informational units, tailored to the specific requirements of the text analysis task at hand \citep{pak2018text}. Recent years have seen the emergence of neural approaches for document and discourse segmentation. \cite{koshorek2018text} suggested hierarchical Bi-LSTMs for document segmentation, \cite{li2018segbot} introduced an attention-based model for both document and discourse segmentation, and \cite{wang2018toward} achieved state-of-the-art results on discourse segmentation using pretrained contextual embeddings. Each segment obtained through text segmentation can be considered as an independent entity within the SNeL framework. This approach would enable the system to process these entities separately, enhancing the granularity of analysis and improving the system's ability to recognize, align, and interpret these entities in response to specific prompts or queries.

In Audio Processing, entities can be conceived as distinct auditory components within an audio stream. This could involve distinct sounds, speech components, or identifiable sound events. The process of Entity Recognition in this domain can include tasks such as sound event detection \citep{mesaros2021sound}, speech recognition \citep{nassif2019speech}, and speaker identification \citep{ye2021deep}. For instance, speaker identification methods aim to recognize an individual entity based on their unique voice characteristics, as explored by \cite{tirumala2017speaker}. Similarly, in the field of sound event detection, algorithms have been developed to recognize entities like particular environmental sounds or musical instruments within an audio clip \citep{purwins2019deep}. The recent advances in deep learning have contributed to significant improvements in these tasks, with CNNs, Recurrent Neural Networks (RNNs), and Transformers being commonly employed for feature extraction and temporal modelling. Each recognized entity, whether it be a specific speaker's voice or a distinct sound event, can be processed individually within our proposed framework, thereby increasing the granularity of analysis and improving the system's ability to interpret and interact with these entities.

\subsection{Cross-Modal Learning}
Humans integrate multiple sensory modes for information processing, facilitated by complex neural networks. To mimic this, artificial intelligence needs to proficiently fuse multi-modal information. In multi-modal research, data from more than one modality, like images, text, video, and audio, is incorporated. While multi-modal systems query one data mode for any modality output, cross-modal systems strictly retrieve information from a different modality \citep{kaur2021comparative}. The process of cross-modal alignment and information retrieval is a key aspect in SNeL's architecture, by allowing it to map textual prompts to entities in non-textual modalities.

In the field of cross-modal alignment, where the objective is to correlate representations from distinct modalities such as text and images, several notable advancements have been made. ImageBERT \citep{qi2020imagebert} extends the BERT \citep{devlin2018bert} model to jointly learn representations for images and text, enabling effective alignment between these two modalities. In a different approach, OpenAI's CLIP (Contrastive Language-Image Pretraining) \citep{radford2021learning} model aligns images and text in a shared latent space by optimizing the similarity of an image and its corresponding text caption, while minimizing the similarity with other captions. Google's ALIGN model \citep{jia2021scaling} employs a dual-encoder architecture, comprising a text encoder and an image encoder that are trained in tandem on a large-scale dataset of text-image pairs. In a similar way, several models for cross-modal alignment between text and audio were proposed. CM-BERT \citep{yang2020cm} extends BERT to audio inputs for sentiment analysis. In a different approach, \cite{mei2022metric} uses metric learning to create joint representations of text and audio for retrieval.



\subsection{Neuro-Symbolic AI}
Neuro-Symbolic AI represents a fusion of symbolic and sub-symbolic (neural) methods, aiming to leverage the strengths of both approaches to overcome their respective limitations. Several significant works have emerged in this field, each proposing unique strategies to integrate symbolic reasoning with neural computation. The Neuro-Symbolic Concept-Learner \citep{mao2019neuro} is a system that learns visual concepts and semantic parsing of sentences, translating input questions into executable programs and executing them on a latent space representation of the scene. DeepProbLog \citep{manhaeve2018deepproblog}, on the other hand, is a Neural Probabilistic Logic Programming system that extends ProbLog to process neural predicates, effectively combining the power of probabilistic logic programming with the learning capacity of neural networks. Logical Neural Networks \citep{riegel2020logical} seek to create a correspondence between neurons and the elements of logical formulas, thereby embedding logical reasoning within the neural computation process. Lastly, Logical Tensor Networks \citep{serafini2016logic} propose a formalism on first-order language, where truth-values are interpreted as feature vectors, enabling the system to integrate deductive reasoning with data-driven machine learning. These works collectively provide a rich landscape for the exploration and advancement of Neuro-Symbolic AI systems.

The system proposed in this work is a manifestation of Neuro-Symbolic AI, a paradigm that synergistically integrates symbolic reasoning and connectionist learning. The symbolic components are embodied in the application of context-free languages, providing a structured, rule-based framework for decision-making and reasoning. On the other hand, the connectionist components are represented by its neural modules, which draw inspiration from advancements in entity detection across multiple modalities and cross-modal alignment.

\section{Language Design}
This Section provides information about the principles and design of the proposed language, by covering the basic data types, as numbers and Booleans, and operations between them, as the arithmetical or logical operators. Additionally, a central component in SNeL's syntax are the prompts, that can be used to refer to entities in the scene using natural language. All these components can be used to form clauses and retrieve the desired information about the scene.



\subsection{Data Types and Operators}
SNeL supports three primary data types: Numbers, Booleans, and Prompts. Each data type can take on certain values and the operations applicable to them vary based on their type. Numbers can be real or integers, booleans can assume the values of true or false and prompts are a key aspect in the language and more details about it will be presented at Section \ref{sec:prompts}. Table \ref{tab:data-types} summarizes the data types in SNeL along with the possible values they can assume.

\begin{table}[h]
    \centering
    \caption{Supported Data Types}
    \begin{tabular}{cl}
        \toprule
        \textbf{Data Type} & \textbf{Values} \\
        \midrule
        \texttt{Number} & Any integer or real number \\
        \texttt{Boolean} & \texttt{true}, \texttt{false} \\
        \texttt{Prompt} & \texttt{[a text to match with entities]} \\
        \bottomrule
    \end{tabular}
    \label{tab:data-types}
\end{table}

The set of operations in SNeL encompasses arithmetic, logical, comparison operations, and mathematical functions. Arithmetic operations include addition, subtraction, multiplication, division, modulus, and exponentiation for the number data type, while logical operators like logical AND, OR, NOT, and XOR manipulate boolean and prompt literals. Comparison operators allow assessments of relative value, such as less than, greater than, less than or equal to, and greater than or equal to, as well as equality checks with equal to and not equal to operators. Defined in infix notation with operators used between operands, these operations adhere to a precedence order, where lower precedence values signify higher priority, assuring consistent and deterministic evaluation of complex expressions.

In addition to the default operators, SNeL also incorporates a library of mathematical functions, broadening the capacity for expressive queries. This collection encompasses a wide range of functions, including those related to rounding operations, logarithmic computations, exponential calculations, and trigonometric evaluations, among others. These functions are invoked in prefix notation, where the function name precedes its arguments within parentheses. The supported operators, their descriptions, their operant types, and precedence levels are summarized in Table \ref{tab:operators}.

\begin{table}[h]
    \centering
    \caption{Combined Table: Supported Operators, Operand Types, and Mathematical Functions}
    \begin{tabular}{cccc}
        \toprule
        \textbf{Operator/Function} & \textbf{Description} & \textbf{Operand Types} & \textbf{Precedence} \\
        \midrule
        \texttt{+} & Addition & \texttt{Number/Prompt} & 3 \\
        \texttt{-} & Subtraction & \texttt{Number/Prompt} & 3 \\
        \texttt{*} & Multiplication & \texttt{Number} & 2 \\
        \texttt{/} & Division & \texttt{Number} & 2 \\
        \texttt{\%} & Modulus & \texttt{Number} & 2 \\
        \texttt{\^} & Exponentiation & \texttt{Number} & 1 \\ \hline
        \texttt{and} & Logical AND & \texttt{Boolean/Prompt} & 2 \\
        \texttt{or} & Logical OR & \texttt{Boolean/Prompt} & 4 \\
        \texttt{not} & Logical NOT & \texttt{Boolean/Prompt} & 1 \\
        \texttt{xor} & Logical XOR & \texttt{Boolean/Prompt} & 3 \\ \hline
        \texttt{==} & Equal to & \texttt{Number/Boolean/Prompt} & 2 \\
        \texttt{!=} & Not equal to & \texttt{Number/Boolean/Prompt} & 2 \\
        \texttt{<} & Less than & \texttt{Number} & 1 \\
        \texttt{<=} & Less than or equal to & \texttt{Number} & 1 \\
        \texttt{>} & Greater than & \texttt{Number} & 1 \\
        \texttt{>=} & Greater than or equal to & \texttt{Number} & 1 \\ \hline
        \texttt{abs} & Absolute value & \texttt{Number} & - \\
        \texttt{max} & Maximum value & \texttt{Number} & - \\
        \texttt{min} & Minimum value & \texttt{Number} & - \\
        \texttt{floor} & Round down & \texttt{Number} & - \\
        \texttt{ceil} & Round up & \texttt{Number} & - \\
        \texttt{round} & Round to nearest integer & \texttt{Number} & - \\
        \texttt{pow} & Raise a number to a power & \texttt{Number} & - \\
        \texttt{sqrt} & Square root & \texttt{Number} & - \\
        \texttt{exp} & Exponential function & \texttt{Number} & - \\
        \texttt{log} & Natural logarithm & \texttt{Number} & - \\
        \texttt{log2} & Base-2 logarithm & \texttt{Number} & - \\
        \texttt{log10} & Base-10 logarithm & \texttt{Number} & - \\
        \texttt{sin} & Sine function & \texttt{Number} & - \\
        \texttt{cos} & Cosine function & \texttt{Number} & - \\
        \texttt{tan} & Tangent function & \texttt{Number} & - \\
        \texttt{asin} & Arcsine function & \texttt{Number} & - \\
        \texttt{acos} & Arccosine function & \texttt{Number} & - \\
        \texttt{atan} & Arctangent function & \texttt{Number} & - \\
        \texttt{mean} & Mean & \texttt{Number} & - \\
        \texttt{std} & Standard deviation & \texttt{Number} & - \\
        \bottomrule
    \end{tabular}
    \label{tab:operators}
\end{table}

Grouping complex expressions using parentheses is a also supported. By employing parentheses, one can establish the desired precedence and grouping of logical and arithmetic operations within their queries. This allows for the creation of complex expressions that combine multiple conditions, attributes, and operators. The use of parentheses enables users to explicitly specify the order in which operations should be evaluated, ensuring the desired outcome and avoiding any ambiguity.

\subsection{Single Prompts}
\label{sec:prompts}
In SNeL, prompts play a pivotal role as they link the abstract language constructs with real-world entities via the underlying neural network models. They can be a simple, single word, or a more complex phrase or sentence that describes or references the scene's entities in natural language. Prompts may specify an entity's attributes, relationships with other entities, or dynamics within the scene. They are represented as a text in natural language between square brackets, e.g.: \texttt{[a person wearing a red shirt]}.

Considering a scene that consists of an image where the $n$ detected objects on it forms the set of entities, $E=\{e_1,e_2,...,e_n\}$, for a given prompt $p$ and an entity $e_i$, the neural modules will predict the alignment score, $s_i=\text{Align}(p,e_i)$ for each entity $e_i$ in the scene. The scores $s_i$ quantifies the degree of alignment between the entities and the prompts, varying between 0 and 1. Furthermore, as the scene, as well as their detected objects, are fixed, we can consider the same list of objects to maintain consistence across multiple prompts. This way, for each prompt $p_j$, a tuple of scores $S_j$ is produced, where $S_j=(s_1,s_2,...,s_n)$.

\begin{figure}[h]
  \centering
  \includegraphics[width=0.45\linewidth]{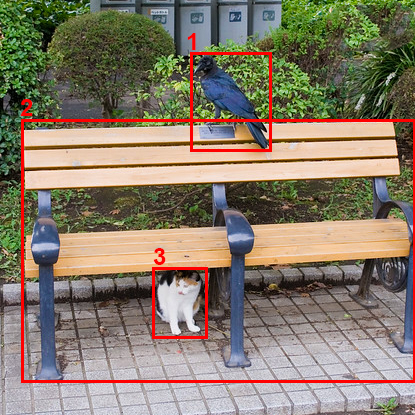}
  \caption{Example of object detection, where each object serves as an entity.}
  \label{fig:entities}
\end{figure}

This process is illustrated through an example in Figure \ref{fig:entities}, where an object detector model detected 3 entities: A bird, a bench, and a cat. Therefore, for each prompt within the query, a tuple of size 3 will be predicted. For example, if the prompt 1 is \texttt{[an animal]}, it is expected that the scores will have greater values on the positions 1 and 3, that refers to the bird and the cat, such as $S_1=(0.8,0.1,0.9)$. If the prompt 2 is \texttt{[an animal that flies]}, it is expected to have values closer to 1 only on the position referring to the bird, such as $S_2=(0.9,0.0,0.2)$.

It's worth noting that the scoring mechanism hinges on the capabilities of the neural network model. The model's design, training data, and level of sophistication can influence the scoring outcomes. Thus, the effectiveness of prompts and the accuracy of scores in representing and querying scenes is closely tied to the underlying neural model's performance.

\subsection{Composite Prompts}
To address for limitations of neural models when the prompts becomes too complex by specifying in very details multiple characteristics and relations, we introduce the concept of composite prompts, which consists of methods for combining multiple prompts through logical and arithmetical operations. This process is evaluated using the scores of the entities for multiple prompts, which are aggregated in high level in a structured way.


In a general case, considering a scene with $n$ entities and two prompts within a given query, $p_a$ and $p_b$, that produces two ordered sequence of scores $S_a=(s^a_1,s^a_2,...,s^a_n)$ and $S_b=(s^b_1,s^b_2,...,s^b_n)$, we can combine $S_a$ and $S_b$ through an arbitrary operation $\odot$ to obtain a new group of scores $S_c=(s^c_1,s^c_2,...,s^c_n)$, where each $s^c_i$ is given by:

\begin{equation}
    s^c_i = s^a_i \odot s^b_i \quad \forall \quad i \in \{1,2,...,n\}
\end{equation}

When grouped operations are performed to produce another list of scores, $S_d=\{s^d_1,s^d_2,...,s^d_n\}$, an operation is applied to the results of a previous operation and another set of scores $s^c_i$ and the type of the operation $\odot$ would depend on the specific grouped operations being performed. as follows:

\begin{equation}
s^d_i = (s^a_i \odot s^b_i) \odot s^c_i \quad \forall \quad i \in \{1,2,...,n\}
\end{equation}

To illustrate, consider the scenario where the scene is an image featuring a diverse fauna. In this case, the detected entities would be the individual animals present within the image. In the case where one might want to filter red birds or animals with fur that are not near a tree, a composite prompt can be used to group those characteristics, as in:

\begin{center}
\begin{BVerbatim}
([a bird] and [a red animal]) or ([an animal with fur] and not [near a tree])
\end{BVerbatim}
\end{center}

Composite prompts allow for more granular and precise entity selection by combining various elementary prompts, while also alleviating issues related to model uncertainty and the lack of explicit reasoning capabilities. By combining multiple prompts, the system can better incorporate context and exploit the inherent structure in visual scenes. This allows for more refined and reliable outputs compared to what would be possible with a single, isolated prompt. More examples of composite prompts are presented in Section \ref{sec:use-cases}.


\subsection{Interpreting Scores}
The interpretation of scores within SNeL is a fundamental aspect of query execution and thus plays a central role in the overall system. As SNeL applies logical operations on these scores to derive final results, the interpretation approach dictates the behavior and results of these operations. We define three approaches for prompt scores interpretation: The Probabilistic, the Fuzzy and the Boolean.

In the Probabilistic interpretation, scores are treated as probabilities. This approach utilizes operations commonly found in probability theory, such as the product rule for "and" operation and complement rule for "not" operation. In the Fuzzy interpretation, scores are treated as degrees of truth ranging from 0 to 1, and standard fuzzy logic operations are employed. Lastly, in the Discrete interpretation, scores are converted into Boolean values, true or false, according to a predetermined threshold, and traditional Boolean logic is employed.

Table \ref{tab:logic-operations-interpretations} provides a summary of how each logical operation is defined under these three different interpretation methods. Each column corresponds to an interpretation approach and depicts the mathematical definition of each logical operation for that approach. The system designer can choose the appropriate interpretation based on the specific requirements of their use-case.

\begin{table}[h]
    \centering
    \caption{Logic Operations between Prompts for Different Interpretation Approaches}
    \begin{tabular}{cccc}
        \toprule
        \textbf{Operator} & \textbf{Probabilistic Interpretation} & \textbf{Fuzzy Interpretation} & \textbf{Boolean Interpretation} \\
        \midrule
        \texttt{and} & $s_a \times s_b$ & $\min(s_a,s_b)$ & $s_a \land s_b$ \\
        \texttt{or} & $s_a + s_b - s_a \times s_b$ & $\max(s_a,s_b)$ & $s_a \lor s_b$ \\
        \texttt{not} & $1 - s_a$ & $1 - s_a$ & $\lnot s_a$ \\
        \texttt{xor} & $s_a \times (1 - s_b) + s_b \times (1 - s_a)$ & $|p_a - p_b|$ & $s_a \oplus s_b$ \\
        \bottomrule
    \end{tabular}
    \label{tab:logic-operations-interpretations}
\end{table}

\subsection{Assembling Clauses}
In the SNeL language, clauses are assembled using a set of predefined keywords that dictate the type of operation to be performed. These keywords include "select", "get", "count", "any", and "all". Each keyword sets the structure for the query, defining what it returns. 

\subsubsection{Select Clause}
The "\texttt{select}" keyword is used to identify entities in a scene that match a given prompt. The output is a list of indices corresponding to the entities on the scene that were selected based on a score threshold. It must be followed by a simple or composite prompt and optional clauses for sorting, ordering and limiting the output. The general structure of a "select" query is: 

\begin{center}
\begin{BVerbatim}
select [entity selection] sort by [sorting attribute] asc/desc limit N
\end{BVerbatim}
\end{center}

The "\texttt{[entity selection]}" prompt specifies the characteristics of the entities that must be selected. "\texttt{sort by}" is an optional clause that sorts the selected entities based on the "\texttt{[sorting attribute]}" prompt, that refers to the characteristics of the attribute used for sorting. The output list can be in ascending or descending order, when using the clauses "\texttt{asc}" or "\texttt{desc}", respectively. In scenarios where no sorting prompt is provided, the "\texttt{asc/desc}" clause can still be employed. However, in these cases, the sorting is conducted based on the scores corresponding to the entity selection prompt. Finally, the \texttt{limit N} clause will limit the number of the entities in the output list by N.

For instance, the query "\texttt{select [a dog] sort by [a dark fur] desc 3}" instructs the system to identify the three dogs possessing the darkest fur from the set of detected entities.

\subsubsection{Get Clause}
The "\texttt{get}" keyword is used to retrieve specific attributes of entities that match a given prompt. The output of a "\texttt{get}" query is a list of attribute values, corresponding to the entities on the scene that match the prompt. It must be followed by an attribute prompt, a "\texttt{from}" clause indicating the target prompt, and optional clauses for sorting, ordering and limiting the output. The general structure of a "\texttt{get}" query is:

\begin{BVerbatim}
get [attribute request] from [entity selection] sort by [sorting attribute] asc/desc
limit N
\end{BVerbatim}

The "\texttt{[attribute request]}" prompt refers to the desired attribute and the "\texttt{[entity selection]}" prompt specifies the entities that should be considered. The "\texttt{sort by [sorting attribute]}", "\texttt{asc/desc}" and \texttt{limit N} clauses works the same way as in the "\texttt{select}" keyword.

As an example, the query "\texttt{get [the color] from [a bird] sort by [a bird close to the tree] desc limit 2}" instructs the system to identify the colors of the two birds closest to the tree from the set of detected entities.

\subsubsection{Count Clause}
The "\texttt{count}" keyword is used to count the number of entities that match a given prompt. Therefore, it returns an interger. The structure of a "\texttt{count}" query is: 

\begin{center}
\begin{BVerbatim}
count [entity selection]
\end{BVerbatim}
\end{center}

Multiple "\texttt{count}" clauses can be used within a single query, since they represent an integer value. Additionally, parenthesis can be used as in function notation to create a more readable query. For example, to return the ratio of red cars in a parking lot, we can use the following query:

\begin{center}
\begin{BVerbatim}
count([a red car] and [a car in a parking lot]) / count([a car in a parking lot])
\end{BVerbatim}
\end{center}

\subsubsection{All and Any Clauses}
The "\texttt{all}" and "\texttt{any}" keywords are used to check if all or any entities match a given prompt, respectively. The structures of "\texttt{all}" and "\texttt{all}" queries are similar to the "count" query: 

\begin{center}
\begin{BVerbatim}
all [entity selection] limit N
any [entity selection] limit N
\end{BVerbatim}
\end{center}

These queries return true or false. "all" query returns true if all entities match the prompt, and the "any" query returns true if at least one entity matches the prompt. As in the query "\texttt{all} [a bird]" checks if all the entities in the scene are birds or in "\texttt{any} [a cat]", that verifies if there is any cat among all entities. The "\texttt{limit}" clause is optional and is used to select N entities in the output. For example, verify that at least one of the three biggest birds are flying, we can use:

\begin{center}
\begin{BVerbatim}
any [a flying bird] sort by [a big bird] desc limit 3
\end{BVerbatim}
\end{center}

Additionally, to verify if the two youngest persons are near the lake, we can use:

\begin{center}
\begin{BVerbatim}
all [a person near the lake] sort by [an old person] asc limit 2
\end{BVerbatim}
\end{center}

The "\texttt{all}" and "\texttt{any}" clauses can also be used within a query among other expressions through the use of logical operators, as they represent a Boolean value. So for example, to check if all of the persons are wearing glasses and at least one of them is sitting on a chair, we can use the following:

\begin{center}
\begin{BVerbatim}
all([a person wearing glasses]) and any([a person sitting on a chair])
\end{BVerbatim}
\end{center}

\section{Language Components}
The main architecture of the implementation of SNeL is composed of two main modules: The Neural Backend and the Language Interpreter. The Neural Backend stands as the sensory unit, responsible for interpreting the scene, discerning entities, associating them with the specified prompts, and predicting attributes when required. It transforms the raw data of the scene into a format primed for semantic processing, thus serving as the perceptive hub of the language. Conversely, the Interpreter operates as the cognitive unit, understanding and processing the semantics of the language through lexical analysis and parsing of queries. In the following Seciont more details about the components of SNeL will be detailed.

\subsection{Neural Backend}
The Neural Backend constitutes the perceptive component of the language. It is a system composed of neural modules that must be capable of executing a few tasks in order to extract the necessary information for the language queries. For handling prompts, it is necessary a detection of the entities of the scene and an alignment with the prompts given in the input query. For the "\texttt{get}" clauses, the Neural Engine must also be capable of predicting the attributes for the entities based on a given prompt. This situation is the case of the operation of Question and Answer (Q\&A) systems, wherein a neural model is tasked with generating an appropriate response to a specified inquiry within a given domain.

In addressing the entity scoring based on the alignment with prompts, the Neural Backend must carry out entity detection and alignment. Regardless of the implementation, the functionality can be represented as a function $f_{ent}: \Omega \rightarrow E$ which extracts all the entities, $E = (e_1, e_2, ..., e_n)$, of the scene $\Omega$, and a function $f_{align}: (E, p_{ent}) \rightarrow S$ which captures the alignment scores, $S = (s_1, s_2, ..., s_n)$, for each entity in a given scene with the entity selection prompt, $p_{ent}$. Each score $s_i$ indicates the degree of alignment of entity $e_i$ with the entity prompt $p_{ent}$, with a value in the range $[0,1]$. This setup allows the Neural Backend to understand the scene, identify and differentiate entities, and associate them with the input prompts in a quantitative manner, laying the foundation for the handling of complex queries.

For handling "\texttt{get}" clauses, the Neural Backend must also predict attributes of the entities based on an attribute prompt $p_{attr}$. This can be denoted as a function $f_{attr}: E \rightarrow A$, where $A$ is the set of possible attribute values. For each entity $e_i$, the function $f_{attr}(e_i)$ predicts the attribute value of $e_i$ based on the attribute prompt $p_{attr}$.




\subsection{Interpreter}
The Interpreter serves as the linguistic component of SNeL. This module is responsible for interpreting the input language queries, which involves transforming the raw input into a structured format that the Neural Backend can interpret and process. The Interpreter achieves this through a two-step procedure: Lexical Analysis and Parsing. The former involves tokenizing the input query into identifiable sequences or lexemes, and the latter is focused on structuring these tokens into a syntactic tree known as a Parse Tree. Together, these two steps allow the Interpreter to process the input query effectively and lay the groundwork for the neural processing carried out by the Neural Backend. This entire process ensures the proper translation of the language query into actions that can be understood and executed by the Neural Backend. The following sections provide a more detailed overview of the Lexical Analysis and Parsing stages involved in the interpretation of a SNeL query.

\subsubsection{Lexical Analysis}
The initial phase in the interpretation of a SNeL query is Lexical Analysis. This stage scans the input query string and transforms it into a series of meaningful sequences, known as lexemes. Each lexeme is classified and associated with a corresponding token, enabling the Interpreter to understand and process the query more effectively. Formally, a lexer, or tokenizer, can be viewed as a function $L: \Sigma^* \rightarrow T^*$, where $\Sigma^*$ is the set of all possible strings over the alphabet of the language, in this case, the set of all possible SNeL queries, and $T^*$ is the sequence of tokens that represent lexemes in the language.

For each token is defined a regular expression to be used to extract it from the input text. The tokens in the SNeL language are broadly divided into several categories, as shown in Table \ref{tab:tokens-categories}.

\begin{table}[h]
    \centering
    \caption{Token Categories and Examples in SNeL Language}
    \begin{tabular}{ll}
        \toprule
        \textbf{Category} & \textbf{Example Tokens} \\
        \midrule
        Literals and Prompts & \texttt{NUMBER, BOOLEAN, PROMPT} \\
        Identifiers & \texttt{ID} \\
        Arithmetic Operators & \texttt{PLUS, MINUS, TIMES, DIVIDE, MODULO, POWER} \\
        Comparison Operators & \texttt{EQ, NE, LT, LE, GT, GE} \\
        Logical Operators & \texttt{AND, OR, NOT, XOR} \\
        Punctuation & \texttt{LPAREN, RPAREN, COMMA} \\
        Reserved Words & \texttt{SELECT, COUNT, GET, ANY, ALL, SAMPLE, SORT\_BY, SORT\_ORDER, LIMIT} \\
        \bottomrule
    \end{tabular}
    \label{tab:tokens-categories}
\end{table}


\subsubsection{Parsing}
The second phase of the Interpreter module is Parsing, which takes the token sequence outputted by the lexical analysis and structures it into a syntactic tree known as a Parse Tree. This Parse Tree represents the syntactic structure of the input query as per the rules defined in the language's grammar and can be formally defined as a function $P: T^* \rightarrow \Gamma$, where $T^*$ is the sequence of tokens from the lexical analysis, and $\Gamma$ is a Parse Tree representing the syntactic structure of the input as per the rules of the SNeL language's grammar.

In the context of the SNeL language, the parsing is guided by a context-free grammar that dictates the valid structures of SNeL queries. For example, a simple rule in the grammar could be that a "\texttt{select}" clause must be followed by an entity prompt and a "\texttt{get}" clause must be followed by an attribute prompt. Therefore, the parser is in charge of checking the syntactic correctness of the input query. If the input does not conform to the grammar rules, the parser will raise an error, and the query will be rejected.





\section{Use Cases and Examples}
In this section, we provide examples of SNeL commands across multiple domains—images, audio, text, and videos—to underscore its broad applicability and versatility. Each domain has a dedicated subsection containing a hypothetical scene and a corresponding table of commands. These examples highlight the expressiveness of SNeL in diverse contexts, demonstrating its utility in facilitating nuanced interactions with AI models across different types of data.

\label{sec:use-cases}
\subsection{Image Domain}

\textbf{Scene description:} An image of a public park filled with people, animals, and various other objects.

\begin{table}[h]
    \centering
    \caption{Examples of SNeL Commands in the Image Domain}
    \renewcommand{\arraystretch}{1.5}
    \begin{tabularx}{\textwidth}{XX}
        \toprule
        \textbf{Command Description} & \textbf{Command} \\
        \midrule
        Return cats near the lake or dogs near the fountain & \texttt{select [cat near the lake] or [dog near the fountain]} \\
        
        Get the youngest person wearing a red shirt & \texttt{select [person with red shirt] order by [a young person] desc limit 1} \\
        
        Calculate the percentage of dogs & \texttt{100 * count [a dog] / count [an animal]} \\
        
        Check if exists a person with a red shirt & \texttt{any [person with red shirt]} \\

        Check if exists at least three persons with glasses & \texttt{count [person wearing glasses] >= 3} \\

        Get the shirt color of the three persons closest to the lake & \texttt{get [color of shirt] from [a person] order by [a person near the lake] desc limit 3} \\
        \bottomrule
    \end{tabularx}
    \label{tab:image-domain-commands}
\end{table}

\subsection{Video Domain}

\textbf{Scene description:} A video recording of a busy city traffic scene, showcasing various vehicles such as cars, trucks, buses, motorcycles, and bicycles, as well as pedestrians crossing streets, traffic signals, and audio cues like honks, sirens etc.

\begin{table}[h]
    \centering
    \caption{Examples of SNeL Commands in the Video Domain}
    \renewcommand{\arraystretch}{1.5}
    \begin{tabularx}{\textwidth}{XX}
        \toprule
        \textbf{Command Description} & \textbf{Command} \\
        \midrule
        Select segments where a pedestrian is jaywalking & \texttt{select [a jaywalking pedestrian]} \\
        
        Get the color of cars that crossed the intersection when the signal was red & \texttt{get [the car color] from [a car crossing on red]} \\
        
        Count the number of motorcycles that passed & \texttt{count [a motorcycle]} \\
        
        Verify if there is any segment where an ambulance is passing & \texttt{any [an ambulance]} \\
        
        Check if all lanes have traffic & \texttt{all [a lane with traffic]} \\
        
        Select 5 segments where a truck is making a turn & \texttt{select [a truck making a turn] limit 5} \\
        
        \bottomrule
    \end{tabularx}
    \label{tab:video-domain-commands}
\end{table}

\subsection{Audio Domain}

\textbf{Scene description:} An audio recording of an orchestral performance, with various instruments such as violins, cellos, flutes, trumpets, and drums being played at different times.

\begin{table}[h]
    \centering
    \caption{Examples of SNeL Commands in the Audio Domain}
    \renewcommand{\arraystretch}{1.5}
    \begin{tabularx}{\textwidth}{XX}
        \toprule
        \textbf{Command Description} & \textbf{Command} \\
        \midrule
        Get audio segments when the violin is playing & \texttt{select [a violin]} \\
        
        Check if there is any solo trumpet performance & \texttt{any [a solo trumpet]} \\
        
        Count the number of times the drums were played & \texttt{count [a drum]} \\
        
        Return the segments when both violins and cellos can be heard together & \texttt{select ([a violin] and [a cello])} \\
        
        Calculate the ratio of segments that a flute is playing across all wind instruments & \texttt{count([a flute]) / count([a wind instrument])} \\

        Get the list of the instruments & \texttt{get [the instrument] from [an instrument playing]} \\
        
        \bottomrule
    \end{tabularx}
    \label{tab:audio-domain-commands}
\end{table}

\subsection{Text Domain}

\textbf{Scene description:} A collection of social media posts containing text and emojis from a public figure's account. 

\begin{table}[h]
    \centering
    \caption{Examples of SNeL Commands in the Text Domain}
    \renewcommand{\arraystretch}{1.5}
    \begin{tabularx}{\textwidth}{XX}
        \toprule
        \textbf{Command Description} & \textbf{Command} \\
        \midrule
        Select posts related to climate change & \texttt{select [a post about climate change]} \\
        
        Check if any post expresses gratitude & \texttt{any [a post expressing gratitude]} \\
        
        Get the list of groups most targeted by hate speech & \texttt{get [the targeted group] from [a post with hate speech against a specific group]} \\
        
        Get the sentiments about the future of posts about artificial intelligence in the optimistic, neutral or pessimist categories & \texttt{get [sentiment about the future in the categories: optimistic, neutral or pessimist] from [a post related to artificial intelligence]} \\
        
        Check if at least 50\% of the posts about sports are about soccer & \texttt{count([a post about soccer])/count([a post about sports]) >= 0.5} \\
        
        Get the genre of the movies of the posts & \texttt{get [the movie genre] from [a post about a movie]} \\
        
        \bottomrule
    \end{tabularx}
    \label{tab:text-domain-commands}
\end{table}

\section{Conclusion and Future Work}
In this paper, we introduced SNeL, a high-level query language model grounded in the ontological concept of entities for reasoning about a scene. This, together with the development of new multimodal deep learning models, allows for the idea to be extended for multiple domains, such as images, audio, video, and text. By integrating neural models with a Context-Free Grammar, our system forms a part of the wider ecosystem of Neuro-Symbolic AI. Through the diverse set of examples provided, we have shown the ability of SNeL to generate precise and granular queries, thereby allowing more efficient and targeted retrieval of information.

However, a key consideration is that the effectiveness of SNeL is deeply contingent on the underlying models used to interpret the data. In essence, it is only as powerful as the models it interacts with. Therefore, the ability to accurately decipher the scene and process the commands in the SNeL language is largely dependent on the quality and capabilities of the models used in the different domains. Moreover, these models must be accurately tuned and trained for the specific tasks and domains for the proposed language to operate at its fullest potential.

Looking forward, our future work will focus on testing SNeL with various models in different domains to explore and evaluate its performance and effectiveness more comprehensively. This will involve identifying the most suitable models in each domain, and training these models to optimize their performance when used in conjunction with SNeL. 

Another promising direction for future work is the expansion of SNeL's language constructs. The addition of more clauses and expressions to the language will enrich its expressiveness and flexibility, thereby broadening its potential application areas. This could include incorporating advanced semantic understanding and reasoning capabilities, which would significantly enhance SNeL's performance in tasks involving complex context understanding.

\clearpage
\bibliographystyle{unsrtnat}
\bibliography{references}  






\end{document}